\documentclass{article}
\usepackage{ijcai24}

\usepackage{times}
\usepackage{soul}
\usepackage{url}
\usepackage[hidelinks]{hyperref}
\usepackage[utf8]{inputenc}
\usepackage[small]{caption}
\usepackage{graphicx}
\usepackage{amsmath}
\usepackage{amsthm}
\usepackage{booktabs}
\usepackage{algorithm}
\usepackage{algorithmic}
\usepackage[switch]{lineno}

\title{An Interactive Human-Machine Learning Interface for Collecting and Learning from Complex Annotations}

\author{
Jonathan Erskine$^1$
\and
Matt Clifford$^1$\and
Alexander Hepburn$^{1}$\and
Raúl Santos-Rodríguez$^{1}$\\
\affiliations
$^1$University of Bristol\\
\emails
\{jonathan.erskine, matt.clifford, alex.hepburn, enrsr\}@brisol.ac.uk
}

\begin{document}

\maketitle

\begin{abstract}
    Human-Computer Interaction has been shown to lead to improvements in machine learning systems by boosting model performance, accelerating learning and building user confidence. In this work, we aim to alleviate the expectation that human annotators adapt to the constraints imposed by traditional labels by allowing for extra flexibility in the form that supervision information is collected. For this, we propose a human-machine learning interface for binary classification tasks which enables human annotators to utilise counterfactual examples to complement standard binary labels as annotations for a dataset. Finally we discuss the challenges in future extensions of this work.
\end{abstract}

\section{Introduction}
% the problem scenario - current landscape
While the field of Artificial intelligence (AI) research has undoubtedly made remarkable progress in recent years, it is important to recognize that this expansion does not necessarily imply a proportionate step towards a form of AI that has ability to understand, learn, and apply knowledge across various domains. Instead, we are witnessing the proliferation of specialized tools; each designed to tackle specific tasks with remarkable proficiency, often with a high associated cost. 

For example, Large Language Models (LLMs) excel in their designated domains \cite{wei2022emergent}, but their exceptional performance is often reliant on the availability of large datasets \cite{devlin-etal-2019-bert} and their capabilities are often limited when confronted with unfamiliar or unforeseen challenges \cite{collins2022structured}. In this paper we ask if the reliance on large amounts of data and poor model generalisation can be alleviated through human-machine interaction, and propose an interface that enables human annotators to evaluate model behaviour and increase the complexity of individual annotations where required.

Many recent approaches in \emph{human-in-the-loop} (HITL) machine learning have focused on providing post-hoc explanations for the decisions generated by complex machine learning models \cite{wang2021putting,jimaging3040056}. By integrating these explanations into the pipeline, researchers aim to empower humans to detect the presence of spurious correlations. Such correlations often imply biases present in the training data, which in turn presents a need for more intelligent data selection.
\emph{Active learning} typically engages humans in the annotation process to strategically choose samples that will most effectively improve the model's performance~\cite{Ren2022-ei}, usually selected to reduce a form of uncertainty~\cite{prince2004does}, mitigating the need for large datasets. However, solely relying on labels for supervision information may be too inflexible for some tasks and may under-utilise the skills of the annotators. 
The field of \emph{Machine Teaching} provides flexibility by allowing humans to select groups of samples that effectively describe concepts and patterns \cite{Zhu2015}. Recent work combines active learning and machine teaching to automatically generate the minimum viable dataset to learn a concept, and then apply an active learning approach to refine this initial prototype \cite{MOSQUEIRAREY2021553}. 

Human involvement in the learning pipeline is not only restricted to data selection. Humans can provide feedback on the explanations produced by models e.g. by adjusting decision boundaries for a 3D segmentation task \cite{jimaging3040056} according to human input, or to augment training examples. Kaushik et al. create a new dataset of counterfactual pairs\cite{Kaushik2020Learning}; a counterfactual is usually defined as a point belonging to the other class which is closest to the current instance. Kaushik et al. generate minimally different counterfactuals by employing human annotators to change film reviews from positive to negative and vice versa while avoiding any unnecessary changes. They demonstrate that models trained with a mixture of original data and human-generated counterfactuals produce robust models that improve generalisation for a sentiment classification task on the IMDB dataset. 

We consider the case where a human expert is familiar with the problem, but no additional data can be generated. Relaxing the method employed by Kaushik, we enable the annotator to indicate the direction from an observation along which we expect a \emph{counterfactual} observation. Through the addition of these annotations, we allow a human annotator to influence the decision boundary of a model during training, without the need for additional data.

\section{Methodology}
\label{methodology}
% Describe the system to be demonstrated, the application domain

\begin{figure*}[t]
%\vskip 0.2in
    \includegraphics[width=\textwidth]{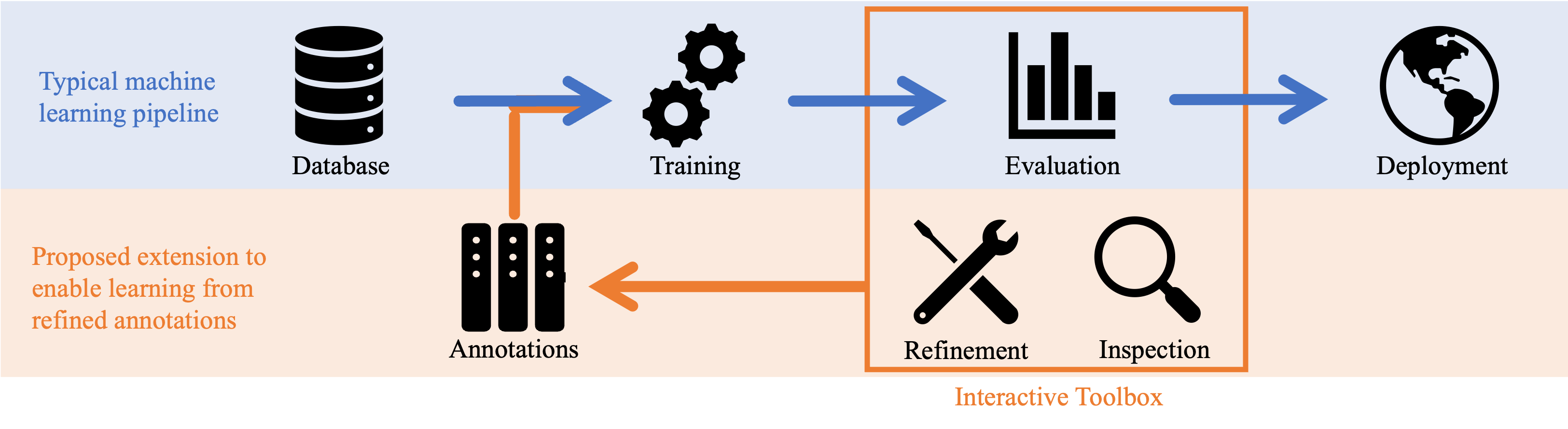}
    \caption{A typical machine learning pipeline contrasted with our proposed method.}
    \label{fig:method-pipeline}
%\vskip -0.2in
\end{figure*}

We demonstrate our approach on a simple synthetic example for which a human expert can provide valuable insights. We describe an interface between human and machine learning system to enable evaluations and machine-intelligible annotations and adapt the training process to learn from these annotations. This extension of the traditional ML pipeline is illustrated in Figure \ref{fig:method-pipeline}.

\subsection{Learning Task}
\label{subsec:learning_task}
We propose 2-dimensional binary classification with separable clusters conforming to typical shapes. As the classes are clearly separable, given any observation, a change in classification can be achieved through a simple translation. Problems of this nature are readily solvable for any human observer, who could consider any instance and indicate whether a positive or negative shift in the either direction would lead to a change in class. 

In traditional supervised learning for classification, we assume a dataset of pairs $\{\mathbf{x_i}, y_i\}$, where $\mathbf{x_i}\in\mathcal{R}^n$ and $y_i$ is the target class label~\cite{hastie2009elements}. Here, let us consider instead that we have access to doublets of the form of an example $\mathbf{x_i}$ and any number of user-defined direction vectors $\mathbf{d_i}\in\mathcal{R}^n$, along each of which we expect to intersect the decision boundary. This can be summarised as $\{\mathbf{x_i}, \mathbf{K_i}=\{\mathbf{d_j},\mathbf{d_{j+1}},...,\mathbf{d_k}\}\}$ where $k$ is the cardinality of the set of counterfactuals defined for any given example in the dataset, $ 0 \leq k < \infty$.

We can define multiple counterfactual directions from one observation, and can repeat this process for any number of observations. A human observer can pause training at any point, and add more counterfactual directions for any observations. Subsequently, model training is resumed, and this process can be repeated indefinitely.

\subsection{Human-Machine Learning Interface}

\begin{figure*}[t!]
%\vskip 0.2in
\begin{center}
\centerline{\includegraphics[width=\textwidth]{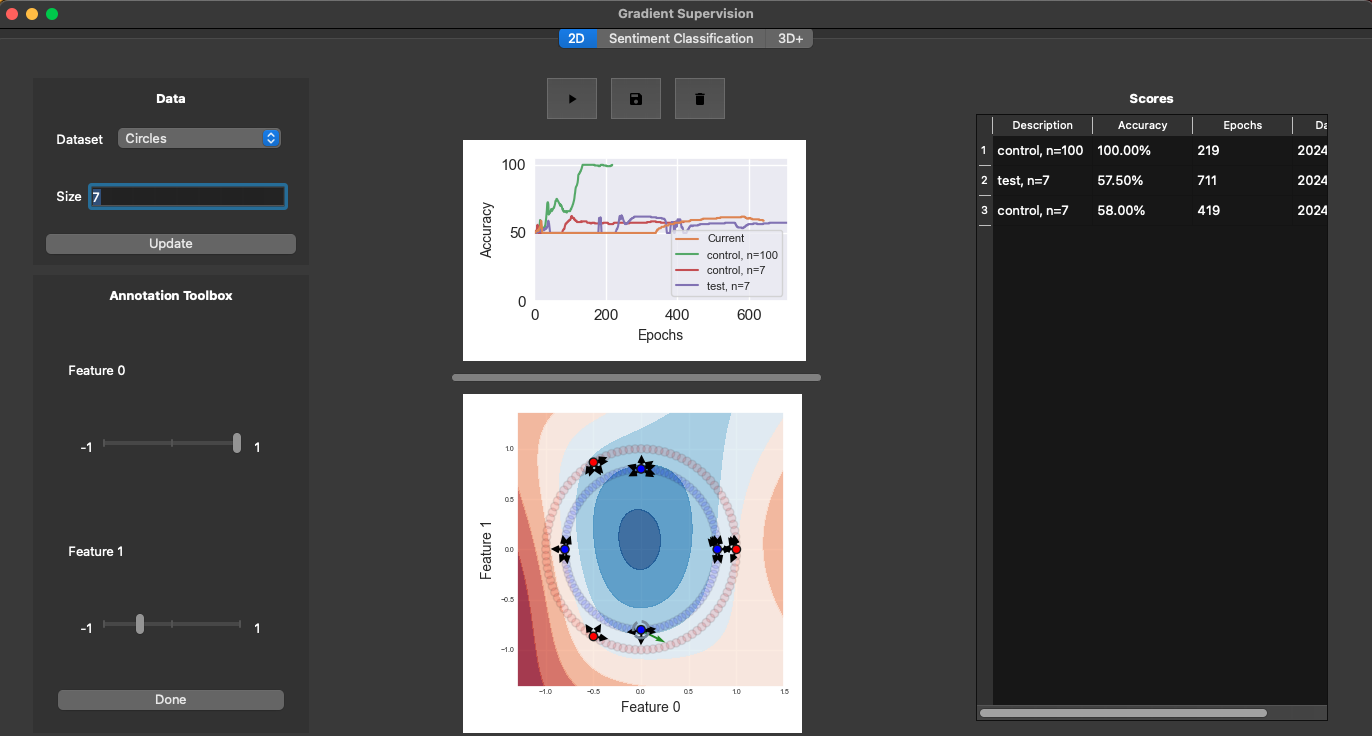}}
\caption{Our Graphical User Interface (GUI) which enables learning from human expert annotations. The current annotation is represented by a green arrow, with prior annotations represented in black. The annotation toolbox (bottom left) enables the annotator to define the direction $\mathbf{d}$ for a selected data point which will lead to a counterfactual observation. The inspection console (bottom-centre) consists of the dataset and the model predicted probabilities across the feature space. The evaluation console (top-centre) enables comparisons across different experiments. This example shows the test accuracy for a control dataset with no annotations and a labelled set which features the counterfactuals directions shown in the inspection console. In both cases the dataset is limited to 9 observations.}
\label{fig:method-gui}

\end{center}
%\vskip -0.2in
\end{figure*}

This annotation scheme defined in Section \ref{subsec:learning_task} requires a graphical user interface, which we illustrate in Figure~\ref{fig:method-gui}. The `Data' section of the interface enables a user to change datasets and experiment with different quantities of observations. The control buttons (top middle) of the GUI enable the user to start and stop training, and save and delete model performance data. 

The center of the console presents the user with a visualisation of the dataset and the test accuracy of the current model and any saved model. The accuracy plots enable an evaluation using a standard measure, as well as comparison of the training for different saved models. The `current' accuracy curve in our example shows the ongoing training effort, with the other series representing previous attempts which have been saved and renamed by the user. Experiments can be added or removed for comparison to allow for experimentation of supervision procedures and multiple human tests at once. Final accuracies are presented in a table on the right hand side of the interface and can be saved to allow for further analysis. 

Under the results we present a live view of the model decision boundaries as they shift during training relative to the labelled training examples (solid) and the test examples (transparent). By observing how the parameter adaptations are affecting the predictive capabilities of the model across multiple iterations, it is possible to draw conclusions about whether the model is effectively and efficiently learning the correct solution. This evaluation, combined with the model accuracy, provides a deeper insight into the learning process. This also allows the user to see in which areas the model is incorrect, and correct it by providing additional counterfactual directions.

The annotation toolbox (lower left) enables a human expert to select any point and generate arrows to indicate counterfactual directions. Existing vectors are shown in black, while the vector being generated is highlighted in green. These vectors, $\mathbf{K}$, must be intelligible to the machine learning pipeline to be used for learning. 

 \subsection{Leveraging Annotations in Learning}
 After defining the counterfactual directions in $\mathbf{K}$, we make the simplifying assumption that the gradient of our predictive model for our given class should be negative along these vectors, to indicate a change in probability. For example, as we move from an observation within the distribution of Class 0 towards the region of Class 1, we expect our probability of being in Class 0 to be decreasing. This component is highly domain-specific; any form of annotation could replace this function, provided the annotation is generated in a form which can be interpreted by an accompanying loss function. The interface is designed such that adapting the annotation toolbox to use a different form of supervision is straight forward. In our demonstration, we satisfy this requirement by developing a novel loss function component which checks the model decision surface and rewards negative gradients in alignment with the human-generated directions. 

\subsubsection{Loss Function}
\label{subsec:loss_fn}

We propose that the gradient of our model $f(\mathbf{x})$ should be aligned with the a human-generated direction $\mathbf{d}_i$ i.e. since the direction is counterfactual, the prediction probability $y_i$ should decrease in this direction. We define a function that aligns the gradients of our model with the direction the human provided, where the sign of the gradient in the given direction is defined as
\begin{equation}
    s_{d_i} = \textrm{sign}(\nabla_{\mathbf{d}_{i}} f(\mathbf{x}_i)),
\end{equation}
where $\nabla_{\mathbf{d}_i}f(\mathbf{x}_i)=\mathbf{d}_i \cdot \nabla f(\mathbf{x}_i)$ is the gradient of the model in the direction of $\mathbf{d}_i$ with respect to input $\mathbf{x}_i$, $f(\mathbf{x}_i)$ is the predicted probability for $\mathbf{x}_i$ belonging to class 0, and the $\textrm{sign}$ function is approximated with $\tanh{(c \nabla_{\mathbf{d}_{i}} f(\mathbf{x}_i) )}$ where we set $c=20$ to create a very steep function. We define our loss function as 
\begin{align}\label{eq:direction_loss}
    \mathcal{L}_{d} &= \frac{1}{N_d} \sum^{N_d}_{i=1} |(2y_i - 1)s_{\mathbf{d}_i} + 1|
    % &\approx\frac{1}{N_d} \sum^N_d_{i=1} \frac{1}{k}\sum^k_{j=1} | \tanh{(c\nabla_{\mathbf{d}_{i}} f(\mathbf{x}_i))} + 1|
\end{align}
where $N_d$ is the number of examples in the training set for which we have labelled directions. This function is $0$ when the gradient of $f(\mathbf{x})$ is in agreement with the human direction, and $2$ otherwise.

We assume an Empirical Risk Minimisation approach with binary cross entropy loss as a baseline (for which a derivation and additional theory is provided by \cite{Goodfellow-et-al-2016}).
%Eq.~\ref{eq:direction_loss} describes our proposed gradient-based loss function.

\section{Demonstration}
% The technology used, the AI techniques involved, the original contribution and innovations of the system, live and interactive aspects.  Indicate links to any supplementary materials. 
The interface is available for download from \href{https://github.com/jmerskine1/interactive_gradients.git}{GitHub}\footnote{\url{https://github.com/jmerskine1/interactive_gradients.git}} with installation instructions and a user manual contained in the \path{README.md} file.

\subsection{Extension to NLP Domain}
We are able to generate direction vectors between observation and counterfactual and utilise these direction vectors for a high-dimensional problem by replicating the counterfactual experiments on the IMDB dataset\cite{Kaushik2020Learning} and utilising our novel loss function. This raises questions around the minimum number of counterfactual observations which can adequately improve model training, for which the user interface may be useful. This would either require some form of active learning component to select those film reviews which are misclassified with the greatest amount of error, or dimensionality reduction which would allow exploration of the feature space in two dimensions.

\appendix

\section*{Ethical Statement}

There are no ethical issues.

\section*{Acknowledgments}
Thanks to UK Research and Innovation (UKRI) and Thales Training \& Simulation Ltd. who are jointly funding Jonathan Erskine's PhD research through the UKRI Centre for Doctoral Training in Interactive Artificial Intelligence under grant EP/S022937/1. This work was partially funded by the UKRI Turing AI Fellowship EP/V024817/1. 

\bibliographystyle{named}
\bibliography{ijcai24}

\begin{thebibliography}{}

\bibitem[\protect\citeauthoryear{Collins \bgroup \em et al.\egroup }{2022}]{collins2022structured}
Katherine~M. Collins, Catherine Wong, Jiahai Feng, Megan Wei, and Joshua~B. Tenenbaum.
\newblock Structured, flexible, and robust: benchmarking and improving large language models towards more human-like behavior in out-of-distribution reasoning tasks, 2022.

\bibitem[\protect\citeauthoryear{Devlin \bgroup \em et al.\egroup }{2019}]{devlin-etal-2019-bert}
Jacob Devlin, Ming-Wei Chang, Kenton Lee, and Kristina Toutanova.
\newblock {BERT}: Pre-training of deep bidirectional transformers for language understanding.
\newblock In {\em Proceedings of the 2019 Conference of the North {A}merican Chapter of the Association for Computational Linguistics: Human Language Technologies, Volume 1 (Long and Short Papers)}, pages 4171--4186, Minneapolis, Minnesota, June 2019. Association for Computational Linguistics.

\bibitem[\protect\citeauthoryear{Goodfellow \bgroup \em et al.\egroup }{2016}]{Goodfellow-et-al-2016}
Ian Goodfellow, Yoshua Bengio, and Aaron Courville.
\newblock {\em Deep Learning}.
\newblock MIT Press, 2016.
\newblock \url{http://www.deeplearningbook.org}.

\bibitem[\protect\citeauthoryear{Hastie \bgroup \em et al.\egroup }{2009}]{hastie2009elements}
Trevor Hastie, Robert Tibshirani, Jerome~H Friedman, and Jerome~H Friedman.
\newblock {\em The elements of statistical learning: data mining, inference, and prediction}, volume~2.
\newblock Springer, 2009.

\bibitem[\protect\citeauthoryear{Kaushik \bgroup \em et al.\egroup }{2020}]{Kaushik2020Learning}
Divyansh Kaushik, Eduard Hovy, and Zachary Lipton.
\newblock Learning the difference that makes a difference with counterfactually-augmented data.
\newblock In {\em International Conference on Learning Representations}, 2020.

\bibitem[\protect\citeauthoryear{Kurzendorfer \bgroup \em et al.\egroup }{2017}]{jimaging3040056}
Tanja Kurzendorfer, Peter Fischer, Negar Mirshahzadeh, Thomas Pohl, Alexander Brost, Stefan Steidl, and Andreas Maier.
\newblock Rapid interactive and intuitive segmentation of 3d medical images using radial basis function interpolation.
\newblock {\em Journal of Imaging}, 3(4), 2017.

\bibitem[\protect\citeauthoryear{Mosqueira-Rey \bgroup \em et al.\egroup }{2021}]{MOSQUEIRAREY2021553}
Eduardo Mosqueira-Rey, David Alonso-Ríos, and Andrés Baamonde-Lozano.
\newblock Integrating iterative machine teaching and active learning into the machine learning loop.
\newblock {\em Procedia Computer Science}, 192:553--562, 2021.
\newblock Knowledge-Based and Intelligent Information \& Engineering Systems: Proceedings of the 25th International Conference KES2021.

\bibitem[\protect\citeauthoryear{Prince}{2004}]{prince2004does}
Michael Prince.
\newblock Does active learning work? a review of the research.
\newblock {\em Journal of engineering education}, 93(3):223--231, 2004.

\bibitem[\protect\citeauthoryear{Ren \bgroup \em et al.\egroup }{2022}]{Ren2022-ei}
Pengzhen Ren, Yun Xiao, Xiaojun Chang, Po-Yao Huang, Zhihui Li, Brij~B Gupta, Xiaojiang Chen, and Xin Wang.
\newblock A survey of deep active learning.
\newblock {\em ACM Comput. Surv.}, 54(9):1--40, December 2022.

\bibitem[\protect\citeauthoryear{Wang \bgroup \em et al.\egroup }{2021}]{wang2021putting}
Zijie~J. Wang, Dongjin Choi, Shenyu Xu, and Diyi Yang.
\newblock Putting humans in the natural language processing loop: A survey, 2021.

\bibitem[\protect\citeauthoryear{Wei \bgroup \em et al.\egroup }{2022}]{wei2022emergent}
Jason Wei, Yi~Tay, Rishi Bommasani, Colin Raffel, Barret Zoph, Sebastian Borgeaud, Dani Yogatama, Maarten Bosma, Denny Zhou, Donald Metzler, Ed~H. Chi, Tatsunori Hashimoto, Oriol Vinyals, Percy Liang, Jeff Dean, and William Fedus.
\newblock Emergent abilities of large language models, 2022.

\bibitem[\protect\citeauthoryear{Zhu}{2015}]{Zhu2015}
Xiaojin Zhu.
\newblock Machine teaching: An inverse problem to machine learning and an approach toward optimal education.
\newblock {\em Proceedings of the {AAAI} Conference on Artificial Intelligence}, 29(1), March 2015.

\end{thebibliography}

\end{document}